\documentclass[conference,10pt]{IEEEtran}

 \usepackage{epsfig}
\usepackage[english]{babel}
\usepackage[latin1]{inputenc}

\usepackage{amssymb}
\usepackage{url}

\usepackage{cite} % Loading the cite package will result in citation numbers being automatically sorted and properly "ranged". i.e., [1], [9], [2], [7], [5], [6] (without using cite.sty) will become: [1], [2], [5]--[7], [9] (using cite.sty)
\usepackage{stfloats}  % Gives LaTeX2e the ability to do double column
\usepackage{amsmath}
\usepackage{subfigure}

\newcommand{\ind}{\mbox{1\hspace{-.25em}l}}
\newcommand{\eN}{\mbox{I\hspace{-.15em}N}}
\newcommand{\eR}{\mbox{I\hspace{-.15em}R}}

\newcommand{\conju}{\mathrm{Conj}}

\newcommand{\DS}{\mathrm{DS}}

\newcommand{\bel}{\mathrm{bel}}
\newcommand{\pl}{\mathrm{pl}}
\newcommand{\betP}{\mathrm{betP}}

\newcommand{\argmax}{\operatornamewithlimits{arg\,max}}

\begin{document}

% paper title
\title{Decision Support with Belief Functions Theory for Seabed Characterization}
% author names and affiliations
% use a multiple column layout for up to three different
% affiliations
\author{\authorblockN{Arnaud Martin}
\authorblockA{ENSIETA, E$^3$I$^2$ - EA3876\\
2, rue Fran{\c c}ois Verny\\
29806 Brest, Cedex 9, France\\
Email: Arnaud.Martin@ensieta.fr}
\and
\authorblockN{Isabelle Quidu}
\authorblockA{ENSIETA, E$^3$I$^2$ - EA3876\\
2, rue Fran{\c c}ois Verny\\
29806 Brest, Cedex 9, France\\
Email: Isabelle.Quidu@ensieta.fr}}

% make the title area
\maketitle

\selectlanguage{english}

\begin{abstract}
The seabed characterization from sonar images is a very hard task because of the produced data and the unknown environment, even for an human expert. In this work we propose an original approach in order to combine binary classifiers arising from different kinds of strategies such as one-versus-one or one-versus-rest, usually used in the SVM-classification. The decision functions coming from these binary classifiers are interpreted in terms of belief functions in order to combine these functions with one of the numerous operators of the belief functions theory. Moreover, this interpretation of the decision function allows us to propose a process of decisions by taking into account the rejected observations too far removed from the learning data, and the imprecise decisions given in unions of classes. This new approach is illustrated and evaluated with a SVM in order to classify the different kinds of sediment on image sonar.
\end{abstract}


% For peer review papers, you can put extra information on the cover
% page as needed:
% \begin{center} \bfseries EDICS Category: 3-BBND \end{center}
%
% for peerreview papers, inserts a page break and creates the second title.
% Will be ignored for other modes.
\IEEEpeerreviewmaketitle
%==================================================================
\section{Introduction}
\label{intro}
\begin{figure}[htb]
\vspace{-0.5cm}
\begin{center}
\includegraphics[height=5cm]{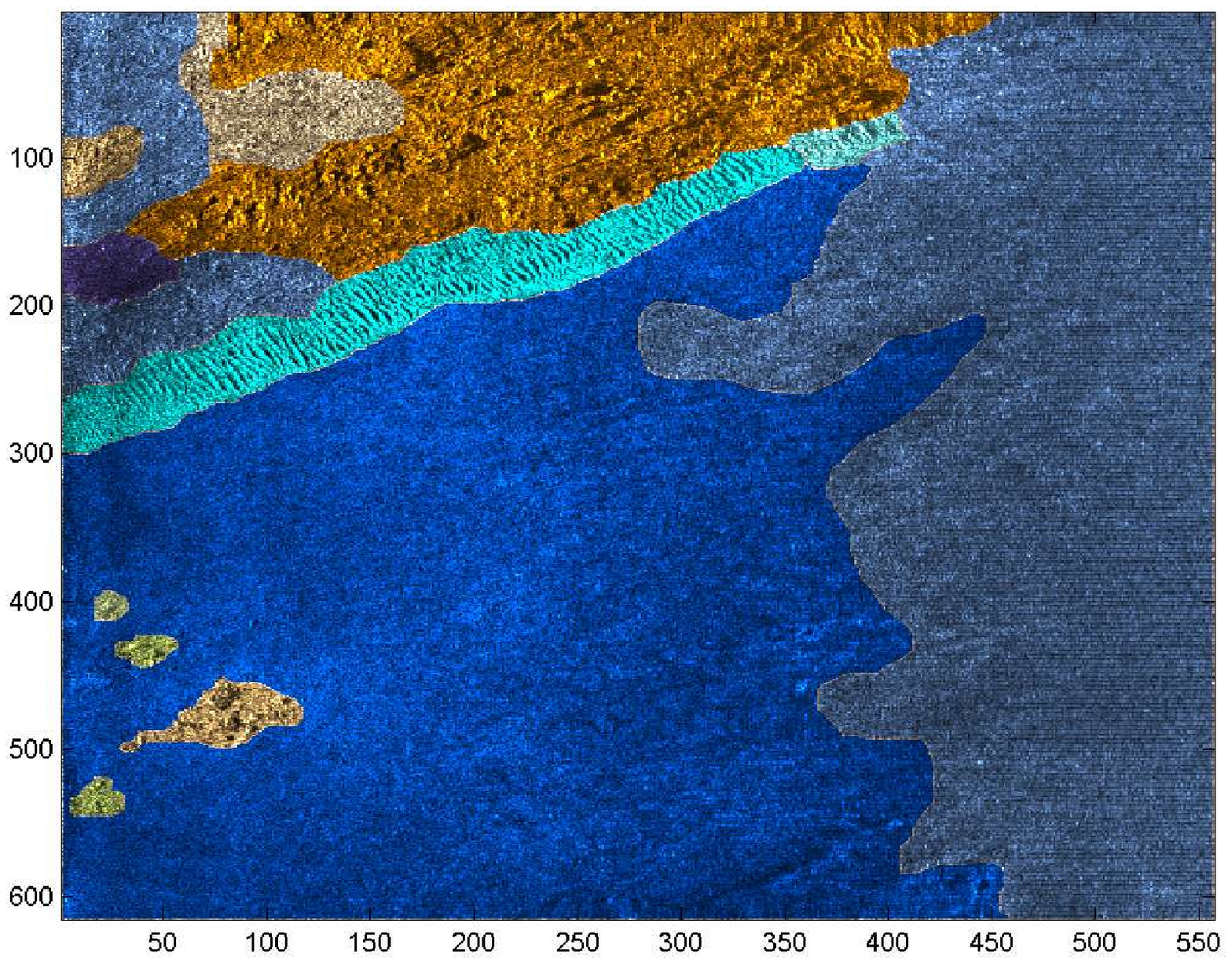}
\includegraphics[height=5cm]{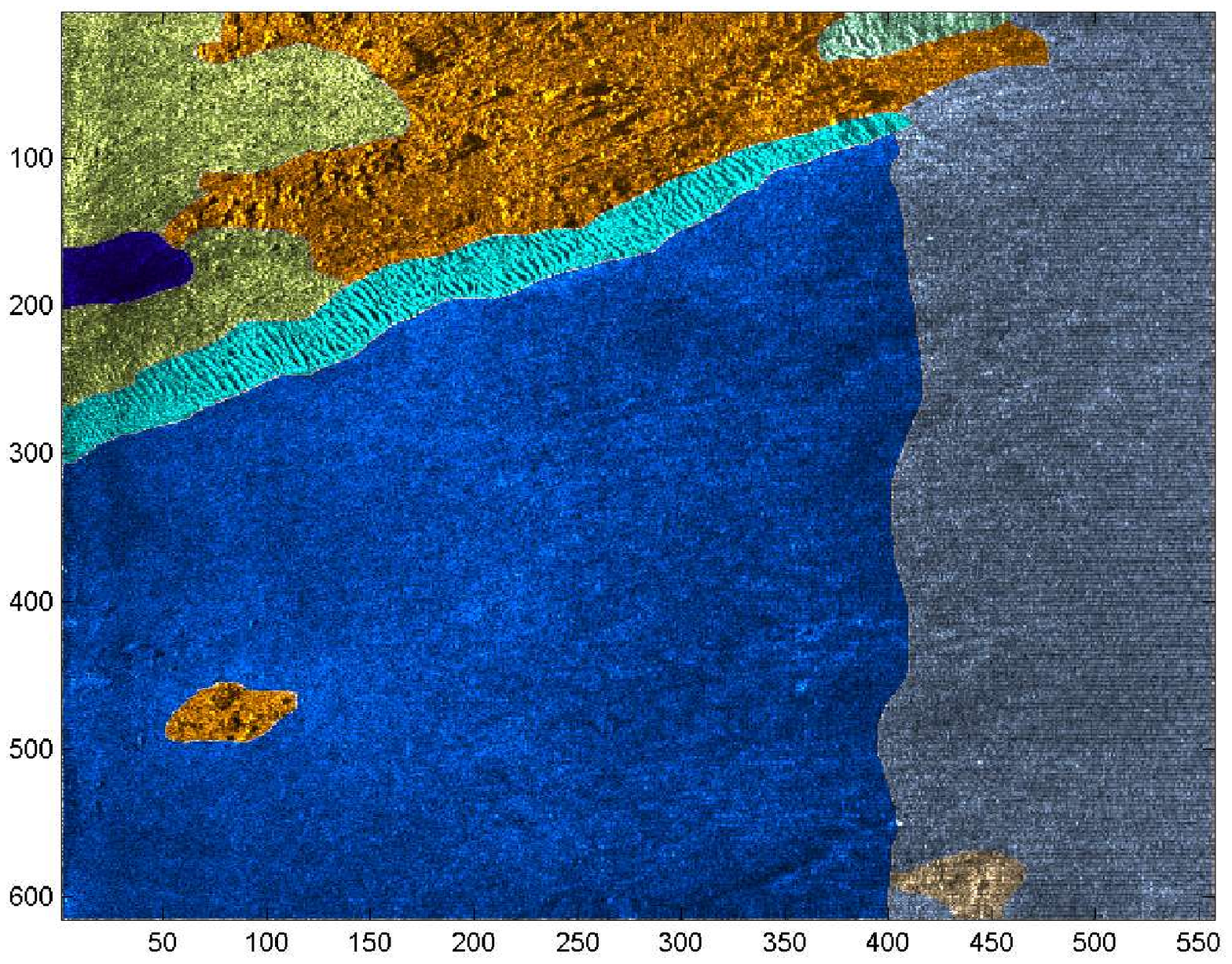}
\end{center}
\vspace{-0.5cm}
\caption{Segmentation given by two experts.}
\vspace{-0.5cm}
\label{expert}
\end{figure}

Sonar images are obtained from temporal measurements made by a lateral, or frontal sonar trailed by the back of a boat. Each emitted signal is reflected on the bottom then received on the antenna of the sonar with an adjustable delayed intensity. Received data are very noisy. There are some interferences due to the signal travelling on multiple paths (reflection on the bottom or surface), due to speckle, and due to fauna and flora. Therefore, sonar images are chraracterized by imprecision and uncertainty; thus sonar image classification is a difficult problem \cite{Martin05}. Figure \ref{expert} shows the differences between the interpretation and the certainty of two sonar experts trying to differentiate types of sediment (rock, cobbles, sand, ripple, silt) or shadow when the information is invisible (each color corresponds to a kind of sediment and the associated certainty of the expert is expressed in terms of sure, moderately sure and not sure) \cite{Martin06}. 

The automatic classification approaches, for sonar images, are based on texture analysis and a classifier such as a SVM \cite{Laanaya06}. The support vector machines (SVM) is based on an optimization approach in order to separate two classes by an hyperplane. For pattern recognition with several classes, this optimization approach is possible (see \cite{Weston98}) but time consuming. Hence a preferable solution is to combine the binary classifiers according to a classical strategy such as one-versus-one or one-versus-rest. The combination of these classifiers is generally formed with very simple approaches such as a voting rule or a maximization of decision function coming from the classifiers. However, many combination operators can be used, especially in the belief functions framework (\emph{cf.} \cite{Martin07b}). Belief functions theory has been already employed in order to combine the binary classifier originally from SVM (see \cite{Aregui07,Quost07}). The operators in the belief functions theory deal with the conflict arising from the binary classifiers. Another interest of this theory is that we can obtain a belief degree on the unions of classes and not only on exclusive classes. Indeed the decisions of the binary classifiers can be difficult to take when data overlap. From the decision function, we can define probabilities in order to combine them (\emph{cf.} \cite{Platt99}). However, a probability measure is an additive measure and so probabilities cannot easily provide a decision on unions of classes unlike belief functions. 

Hence, once the binary classifiers have been combined, we propose belief functions in order to take the decision for one class only if this class is credible enough, for the union of two or more classes otherwise. Moreover, according to the application it could be interesting to not take the decision on one of the learning classes, and reject data too far from the learning classes. Many classical approaches are possible in pattern recognition for outliers rejection (see \cite{Frelicot98,Liu02}). We propose here to integrate outliers rejection in our decision process based on belief functions.

In addition to this new decision process, the originality of the paper concerns the modelization that we propose, \emph{i.e.} how to define the belief functions, on the basis of decision functions coming from the binary classifiers. 

This paper is organized as follows: in section \ref{SVM} we recall the principle of the support vector machines for classification. Next, we present the belief functions theory in section \ref{belief} in order to propose in section \ref{beliefSVM} our belief approach to combine the binary classifiers and to provide a decision process allowing the outliers rejection and the indecision expressed as possible decisions on unions. This approach is evaluated for seabed characterization on sonar images in section \ref{application}.

\section{SVM for classification}
\label{SVM}

Support vector machines were introduced by \cite{Vapnik98} based on the statistical learning theory. Hence, SVM can be used for estimation, regression or pattern recognition like in this paper.

\subsection{Principle of the SVM}

The support vector machine approach is a binary classification method. It classifies positive and negative patterns by searching the optimal hyperplane that separates the two classes, while guaranteeing a maximum distance between the nearest positive and negative patterns. The hyperplane that maximizes this distance called \textit{margin} is determined by particular patterns called support vectors situated at the bounds of the margin. These only few support vector numbers are used to classify a new pattern, which makes SVM very fast. The power of SVM is also due to their simplicity of implementation and to solid theoretical bases.

If the patterns are linearly separable, we search the hyperplane $y = w.x + b$ which maximizes the margin between the two classes where $w.x$ is the dot product of $w$ and $x$, and $b$ the bias. Thus $w$ is the solution of the convex optimization problem:
\begin{equation}
\text{Min} \quad\|w\|^2/2
\end{equation}
subject to:
\begin{equation}
\label{con}
\quad y_t(w.x_t+b)-1\geq 0 \quad\forall t=1,\dots, l, 
\end{equation}
where $x_t \in \eR^d$ stands for one of the $l$ learning data, and \linebreak $y_t \in \{-1,+1\}$ the associated class. We can solve this optimization problem with the following Lagrangian:
\begin{equation}
\displaystyle L=\frac{\|w\|^2}{2} \sum_{t=1}^l \Lambda_t \left(y_t(w.x_t-b)-1\right),
\end{equation}
where the $\Lambda_t \geq 0$ are the Lagrange multipliers, satisfying $\displaystyle \sum_{t=1}^l \Lambda_t y_t=0$.

If the data are not linearly separable, the constraints (\ref{con}) are relaxed with the introduction of positive terms $\xi_t$. In this case we search to minimize:
\begin{equation}
\frac{1}{2}\parallel w\parallel^2+C\sum_{t=1}^{l}\xi_t,
\end{equation}
with the constraints given for all $t$:
\begin{equation}\label{eq8}
\left\{\begin{array}{l}
y_t(w.x_t+b)\geq 1-\xi_t\\
\xi_t\geq 0
\end{array}\right.
\end{equation}
where $C$ is a constant given by the user in order to weight the error. This problem is solved in the same way as the linear separable case with Lagrange multipliers $0 \leq \Lambda_t \leq C$.

To classify a new pattern $x$ we simply need to study the sign of the decision function given by:
\begin{equation}
\label{f_i1}
f(x) = \sum_{t\in SV}y_t\Lambda_t x_t.x - b,
\end{equation}
where $SV = \{t~; \Lambda_t > 0\}$ for the separable case and $SV = \{t~; 0<\Lambda_t < C\}$ for the non-separable case, is the set of indices of the support vectors, and $\Lambda_t$ are the Lagrange multipliers.

In the nonlinear cases, the common idea of the kernel approaches is to map the data in a high dimension. To do that we use a kernel function that must be bilinear, symmetric and positive and corresponds to a dot product in the new space. The classification of a new pattern $x$ is given by the sign of the decision function:
\begin{eqnarray}
\label{f_i2}
f(x)  = \sum_{t\in SV}y_t\Lambda_tK(x,x_t) - b
\end{eqnarray}
where $K$ is the kernel function. The most used kernels are the polynomial $K(x,x_t)=(x.x_t + 1)^\delta,$ $\delta \in \eN$, and the radial basis functions $K(x,x_t) = e^{-\gamma\|x - x_t\|^2}$, $\gamma \in \eR^{+}$. The choice of the kernel is not always easy and generally left to the user. 

\subsection{Multi-class classification with SVM}
We can distinguish two kinds of approaches in order to use SVM for classification with $n$ classes, $n>2$. The first one consists in fusing several binary classifiers given by the SVM - the obtained results by each classifier are combined to produce a final result following strategies such as one-versus-one or one-versus-rest. The second one consists in considering the optimization problem.

\begin{itemize}
\item Direct approach: in \cite{Weston98}, the notion of margin is extended to the multi-class problem. However, this approach becomes very time consuming, especially in the nonlinear case.
 						
\item one-versus-rest: This approach consists in learning $n$ decision functions $f_i$, $i=1,...,n$ given by the equations (\ref{f_i1}) or (\ref{f_i2}) according to the cases, allowing the discrimination of each class from the $n-1$ others. The affection of a class $w_k$ to a new pattern $x$ is generally given by the relation: $\displaystyle k=\argmax_{i=1,...,n}f_i(x)$. In the nonlinear case, we have to be careful of the parameters of the kernel functions that could have some different orders following the learning binary classifiers. So, it could be better to decide on normalized functions calculated from the decision functions (see \cite{Milgram06,Liu05}). 
 								
\item one-versus-one: Instead of learning $n$ decision functions, we try here to discriminate each class from each other. Hence we have to learn $n(n -1)/2$ decision functions, still given by equations (\ref{f_i1}) or (\ref{f_i2}) according to the different cases. Each decision function is considered as a vote in order to classify a new pattern $x$. The class of $x$ is given by the majority voting rule.

\end{itemize}

Some other methods have been proposed based on previous ones:

\begin{itemize}
\item \textit{Error-Correcting Output Codes} (ECOC): let $w_i$, \linebreak $i=1,...,n$, be the classes, $S_j$, $j= 1,...,s$, the different classifiers ($s=n$ in the case one-versus-rest and \linebreak $s=n(n-1)/2$ in the case one-versus-one), $(M_{ij})$, the matrix of the codes with the classes in row and the classifiers in column, stands for the contribution of each classifier to the final result of the classification (based on the error of all the classifiers). The final decision is given comparing the results of the classifiers with each row of the matrix; the class of a new pattern $x$ is the class giving the least error (see \cite{Dietterich95}).

\item According to the decision functions, \cite{Platt99} defined a probability \eqref{ProbaPlatt} in order to normalize the decision functions. Hence, we can combine the binary classifiers (for both one-versus-rest and one-versus-one cases) with a Bayesian rule (see \cite{Lauberts06}) or with more simple rules (see \cite{Quost07}).

\item DAGSVM (Directed Acyclic Graph SVM) proposed by \cite{Platt00}: In this approach, the learning is made as the one-versus-one with the learning of $n(n -1)/2$ binary decision functions. In order to generalize, a binary decision tree is considered where each node stands for a binary classifier and each leaf stands for a class.  Each binary classifier eliminates a class and the class of a new pattern is the class given by the last node.

\end{itemize}

\section{Belief functions theory}
\label{belief}

The belief functions theory, also called evidence theory or Dempster-Shafer theory (see \cite{Dempster67, Shafer76}) is more and more employed in order to take into account the uncertainties and imprecisions in pattern recognition. The belief functions framework is based on the use of functions defined on the power set $2^\Theta$ (the set of all the subsets of $\Theta$), where \linebreak $\Theta=\{w_1,\ldots,w_n\}$ is the set of exclusive and exhaustive classes. These \textit{belief functions} or \textit{basic belief assignments}, $m_j$ are defined by the mapping of the power set $2^\Theta$ onto $[0,1]$ with generally:
\begin{equation}
\label{close}
m_j(\emptyset)=0,
\end{equation}
and 
\begin{eqnarray}
\label{hyp1_fonction_masse}
\sum_{X \in 2^\Theta} m_j(X)=1,
\end{eqnarray}
where $m_j(.)$ is the basic belief assignments for an expert (or a binary classifier) $S_j$, $j=1,...,s$. Thus in the one-versus-rest case $s=n$ and in the one-versus-one case $s=n(n-1)/2$.

The equation (\ref{close}) makes the assumption of a closed world (that means that all the classes are exhaustive) \cite{Shafer76}. We can define the belief functions only with:
\begin{equation}
\label{open}
m_j(\emptyset)\geq 0,
\end{equation}
and the world is open (\emph{cf.} \cite{Smets90b}). But in order to change an open world to a closed world, we can add one element in the discriminating space and this element can be considered as the garbage class. The difficulty, as we will see later, is the mass that we have to allocate to this element.

We have two advantages with the belief functions theory compared to the probabilities and Bayesian approaches. The first one is the possibility for one expert (\emph{i.e.} a binary classifier) to decide that a new pattern belongs to the union of some classes without needing to decide an unique class. The basic belief functions are not additive that gives more freedom for the modelization of some problems. The second one is the modelization of some problems without any \textit{a priori} by giving the mass of belief on the ignorances (\emph{i.e.} the unions of classes).

These simple conditions in equations (\ref{close}) and (\ref{hyp1_fonction_masse}), give a large panel of definitions of the belief functions, which is one of the difficulties of the theory. From these basic belief assignments, other belief functions can be defined such as credibility and plausibility. The credibility represents the intensity that the information given by one expert supports an element of $2^\Theta$, this is a minimal belief function given for all  $X \in 2^\Theta$ by:
\begin{eqnarray}
\bel(X)=\sum_{Y \subseteq X, Y \neq \emptyset} m_j(Y).
\end{eqnarray}
The plausibility represents the intensity with which there is no doubt on one element. This function is given for all $X \in 2^\Theta$ by:
\begin{eqnarray}
\begin{array}{rcl}
\pl(X)&=&\displaystyle \sum_{Y \in 2^\Theta, Y\cap X \neq \emptyset} m_j(Y)\\
&=&\bel(\Theta)-\bel(X^c)\\
&=&1-m_j(\emptyset)-\bel(X^c),
\end{array}
\end{eqnarray}
where $X^c$ is the complementary of $X$ in $\Theta$. 

To keep a maximum of information, it is preferable to combine information given by the basic belief assignments into a new basic belief assignment and take the decision on one of the obtained belief functions. Many combination rules have been proposed. The conjunctive rule proposed by \cite{Smets90a} allows us to stay in an open world. It is defined for $s$ experts, and for $X \in 2^\Theta$ by:
\begin{eqnarray}
\label{conjunctive}
m_\conju(X)=\sum_{Y_1 \cap ... \cap Y_s = X} \prod_{j=1}^s m_j(Y_j),
\end{eqnarray}
where $Y_j \in 2^\Theta$ is the response of the expert $j$, and $m_j(Y_j)$ the corresponding basic belief assignment.

Initially, \cite{Dempster67} and \cite{Shafer76} have proposed a conjunctive normalized rule, in order to stay in a closed world. This rule is defined for $s$ classifiers, for all $X \in 2^\Theta$, $X\neq \emptyset$ by:
 \begin{eqnarray}
\label{DS}
\begin{array}{rcl}
m_\DS(X)&=&\displaystyle \frac{1}{1-m_\conju(\emptyset)} \sum_{Y_1 \cap ... \cap Y_s = X}  \prod_{j=1}^s m_j(Y_j)\\
&=& \displaystyle \frac{m_\conju(X)}{1-m_\conju(\emptyset)},
\end{array}
\end{eqnarray}
where $Y_j \in 2^\Theta$ is the response of the expert $j$, and $m_j(Y_j)$ the corresponding basic belief assignment. $m_\conju(\emptyset)$ is generally interpreted as a conflict measure or more exactly as the inconsistence of the fusion - because of the nonidempotence of the rule. This rule applied on basic belief assignments where the only focal elements are the classes $w_j$ (\emph{i.e.} some probabilities) is equivalent to a Bayesian approach. A short review of all the combination rules in the belief functions framework and a number of new rules are given in \cite{Martin07b}.

If the credibility function provides a pessimistic decision, the plausibility function is often too optimistic. The pignistic probability \cite{Smets90b} is generally considered as a compromise. It is calculated from a basic belief assignment $m$ for all $X \in 2^\Theta$, with $X \neq \emptyset$ by:
\begin{eqnarray}
\label{pignistic}
\betP(X)=\sum_{Y \in 2^\Theta, Y \neq \emptyset} \frac{|X \cap Y|}{|Y|} \frac{m(Y)}{1-m(\emptyset)},
\end{eqnarray}
where $|X|$ is the cardinality of $X$.

In this paper, we wish to reject part of the data that we do not consider in the learning classes. Hence a pessimistic decision as to the maximum of the credibility function is preferable. Another criterion proposed by \cite{le_hegarat97}, consists in attributing the class $w_k$ for a new pattern $x$ if:
\begin{eqnarray}
\label{maxBelRejet}
\left\{
\begin{array}{l}
	\bel(w_k)(x)=\displaystyle \max_{1\leq i \leq n} \bel(w_i)(x),\\
	\bel(w_k)(x) \geq \bel(w_k^c)(x).\\
\end{array}
\right.
\end{eqnarray}
The addition of this second condition on the maximum of credibility, allows a decision only if it is nonambiguous, \emph{i.e.} if we believe more in the class $w_k$ than in the subset of the other classes (the complementary of the class). 

Another approach proposed in \cite{Appriou05} considers the plausibility functions and gives the possibility to decide whichever element of $2^\Theta$ and not only the singletons as previously. Thus the new pattern $x$ belongs to the element $A$ of $2^\Theta$ if:
\begin{eqnarray}
\label{DecAppriou}
	A=\argmax_{X \in 2^\Theta} \left(m_b(X)(x)\pl(X)(x)\right),
\end{eqnarray}
where $m_b$ is a basic belief assignment given by:
\begin{eqnarray}
m_b(X)=K_b \lambda_X \left(\frac{1}{|X|^r}\right),
\end{eqnarray}
$r$ is a parameter in $[0,1]$ allowing a decision from a simple class ($r=1$) until the total indecision $\Theta$ ($r=0$). $\lambda_X$ allows the integration of the lack of knowledge on one of the elements $X$ in $2^\Theta$. In this paper, we will chose $\lambda_X=1$. The constant $K_b$ is the normalization factor giving by the condition of the equation (\ref{hyp1_fonction_masse}).

\section{Belief functions theory for classification with support vector machines}
\label{beliefSVM}

In the previous sections, we have described the two main strategies in order to build a multi-class classifier from binary classifiers: the one-versus-rest and one-versus-one approaches. Most of the time the formalism to combine the binary classifier results is different according to the strategy. \cite{Laanaya06b} have proposed a combination approach of the binary classifier decisions based on the belief functions theory given an unique formalism for both one-versus-one and one-versus-rest strategies. The basic belief assignments are defined from confusion matrices of the binary classifiers. Working directly on the classifier decisions allows a loss of information contained first in the decision functions. Thus it could be better to define the basic belief assignments from the decision functions rather than from the confusion matrices (\emph{i.e.} form the classifier decisions). 

However, the decision functions are not normalized, so we can have problems in the combination of this function especially with the one-versus-rest strategy. \cite{Platt99} has defined a probability from these decision functions $f$ such as:
\begin{eqnarray}
\label{ProbaPlatt}
P(y=1/f)=\frac{1}{1+\exp(Af+B)},
\end{eqnarray}
where $A$ and $B$ are calculated in order to get \linebreak $P(y=1/f=0)=0.5$. Different approaches have been proposed for the estimation of these parameters (see \cite{Lin08}). 

\cite{Quost07} uses a one class SVM, introduced by \cite{Scholkopf99}. So the combination can be done only with a one-versus-rest strategy. The decision functions coming from this particular classifier are employed to define some plausibility functions on the singleton $w_i$:
\begin{eqnarray}
\pl(w_i)(x)=\frac{f_i(x)+\rho}{\rho},
\end{eqnarray}
where $f_i(x)$ is the decision function giving the distance between $x$ and the fronter of class $w_i$ and $\rho$ is a factor estimated in the one-SVM algorithm that depends on the kernel ({\em cf.} \cite{Scholkopf99}).

\begin{figure}[htb]
\includegraphics[height=5.3cm]{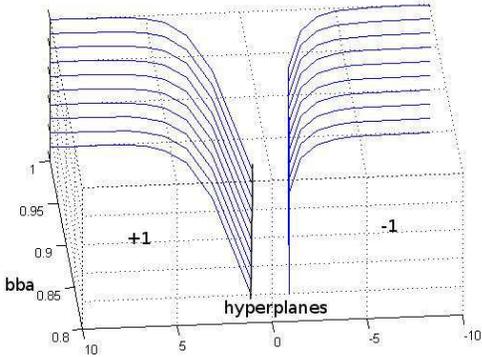}
\vspace{-0.5cm}
\caption{Illustration of the basic belief assignment based on the cumulative density function of the exponential distribution.}
\label{massIllustration}
\end{figure}

The first originality of this paper resides in the definition of the basic belief assignments that we obtain directly from the decision functions $f$ given by the equations (\ref{f_i1}) or (\ref{f_i2}). The basic idea consists in considering the data dispersion in one of the semi-spaces given by the hyperplane, following an exponential distribution. This distribution gives a dispersion of the data around the mean more or less near to the hyperplane, with the opportunity to observe data very far away from the hyperplane. Doing this we keep the basic idea of the SVM. Hence, according to the sign of the decision function (\emph{i.e.} the semi-space defined by the hyperplane), the belief can be obtained by the cumulative density function of the exponential distribution (see figure \ref{massIllustration}). We define the basic belief assignment by:
\begin{eqnarray*}
\left\{
\begin{array}{rcl}
\!\!\!\!m_{i} (w_i)(x)&\!\!\!\!=& \!\!\!\!\!\! \alpha_{i} \left( (1-\frac{1}{2}\exp(-\frac{1}{\lambda_{i,p}} f_i(x))) \ind_{[0,+\infty[}(f_i(x)) \right.\\
 & & \left. \exp(-\frac{1}{\lambda_{i,n}}f_i(x))\ind_{]-\infty,0[}(f_i(x)) \right)\\
\\
\!\!\!\!m_{i} (w_i^c)(x)&\!\!\!\!=&\!\!\!\!\!\! \alpha_{i} \left( \exp(-\frac{1}{\lambda_{i,p}}f_i(x))\ind_{[0,+\infty[}(f_i(x)) \right.\\
& & \!\!\!\!\!\!\!\! \left. (1-\frac{1}{2}\exp(-\frac{1}{\lambda_{i,n}} f_i(x))) \ind_{]-\infty,0(}(f_i(x)) \right)\\
\\
\!\!\!\!m_{i}(\Theta)(x)&\!\!\!\!=&\!\!\!\!\!\!1-\alpha_{i} \\
\end{array}
\right.
\end{eqnarray*}
where $\alpha_{i}$ is a discounting factor of the basic belief assignment, $\lambda_{i,p}$ and $\lambda_{i,n}$ are some parameters depending on the decision functions of class $w_i$ that we define in equation \eqref{lambda}. The ratio $\frac{1}{2}$ is introduced to increase the belief to the class related to the semi-space where the data are located (see figure \ref{massIllustration}). There are many ways to choose or to calculate the discounting factor that is generally close to one. \cite{Elouedi04} proposes a method to obtain the discounting factor that optimizes the decision taking advantage of the pignistic probability. We propose here to calculate this discounting factor according to the good classification rate of binary classifiers. The good classification rates are calculated with the study of the sign of the decision function $f_i$ on the learning data used to determine the model of binary classifiers.

We propose to estimate the $\lambda_i$ parameters from the mean of the decision functions on the learning data in order to be coherent with the exponential distribution. Hence $\lambda_{i,p}$ and $\lambda_{i,n}$ are given by:
\begin{eqnarray}
\label{lambda}
\left\{
\begin{array}{l}
\lambda_{i,p}=\displaystyle \frac{1}{l} \sum_{t=1}^{l} f_i(x)\ind_{[0,+\infty[}(f_i(x)),\\
\lambda_{i,n}=\displaystyle \frac{1}{l} \sum_{t=1}^{l} f_i(x)\ind_{]-\infty,0[}(f_i(x)).
\end{array}
\right.
\end{eqnarray}

This proposed basic belief assignment model allows a good modelization of the information given by the binary classifiers in order to combine them by both one-versus-rest and one-versus-one strategies. Thus for a one-versus-rest strategy, $w_i^c$ represents the union of the other classes than $w_i$, {\em i.e.} \linebreak $\Theta \smallsetminus \{w_i\}$. In the one-versus-one case, the decision functions $f_i$, $i=1,...,n(n -1)/2$ can be rewritten as $f_{ij}$ with $i < j$ and $i,j=1,...,n$, where $i$ and $j$ correspond to the considered classes $w_i$ and $w_j$. In this one-versus-one case, $w_i^c$ must be seen as $w_j$ and the basic belief assignment are given by:
\begin{eqnarray*}
\left\{
\begin{array}{rcl}
\!\!\!\!m_{ij} (w_i)(x)&\!\!\!\!=& \!\!\!\!\!\!  \alpha_{ij}\! \left((1-\exp(-\frac{1}{\lambda_{ij,p}} f_{ij}(x)))\ind_{[0,+\infty[}(f_{ij}(x)) \right. \\
&& \left. + \exp(-\frac{1}{\lambda_{ij,n}} f_{ij}(x)) \ind_{]-\infty,0[}(f_{ij}(x))\right) \\
\\
\!\!\!\!m_{ij} (w_j)(x)&\!\!\!\!=& \!\!\!\!\!\! \alpha_{ij}\! \left(\exp(-\frac{1}{\lambda_{ij,p}} f_{ij}(x))) \ind_{[0,+\infty[}(f_{ij}(x)) \right.\\
&&\!\!\!\! \left. (1-\exp(-\frac{1}{\lambda_{ij,n}} f_{ij}(x))) \ind_{]-\infty,0[}(f_{ij}(x)) \right)\\
\\
\!\!\!\!m_{ij}(\Theta)(x)&\!\!\!\!=& \!\!\!\!\!\! 1-\alpha_{ij} \\
\end{array}
\right.
\end{eqnarray*}
with
\begin{eqnarray}
\left\{
\begin{array}{l}
\lambda_{ij,p}=\displaystyle \frac{1}{l} \sum_{t=1}^{l} f_{ij}(x) \ind_{[0,+\infty[}(f_{ij}(x)),\\
\lambda_{ij,n}=\displaystyle \frac{1}{l} \sum_{t=1}^{l} f_{ij}(x) \ind_{]-\infty,0[}(f_{ij}(x)).
\end{array}
\right.
\end{eqnarray}

We use here the conjunctive normalized rule (equation \eqref{DS}). Thus we can apply this rule in order to combine the $n$ basic belief assignments in the one-versus-rest case and the $n(n-1)/2$ basic belief assignments in the one-versus-one case. When the data overlap a lot, more complicated rules such as proposed in \cite{Martin07b} could be preferred.

For the decision step, we want to keep the possibility to take the decision on a union of classes ({\em i.e.} when we can not decide between two particular classes) and also to not take a decision when our belief in one focal element is too weak. Thus we propose the following decision rule in two steps:
\begin{enumerate}
\item The decision rule of the maximum of the credibility with reject defined by the equation \eqref{maxBelRejet} is applied in order to determine the patterns that do not belong to the learning classes.
\item The decision rule given by the equation \eqref{DecAppriou} is next applied to the non-rejected patterns.
\end{enumerate}

Another possible decision process could be first the application of the decision rule given by the equation (\ref{DecAppriou}), and next the decision rule of the maximum of the credibility with reject on the imprecise patterns that first belong to the unions of classes. On the illustrated data given in the next section, we obtain similar results. We call this decision process (2-1) and the previous one (1-2).

\section{Application}
\label{application}

\subsection{Sonar data}  
Our database contains 42 sonar images provided by the GESMA (Groupe d'Etudes Sous-Marines de l'Atlantique). These images were obtained with a Klein 5400 lateral sonar with a resolution of 20 to 30 cm in azimuth and 3 cm in range. The sea-bottom depth was between 15 m and 40 m.

Some experts have manually segmented these images giving the kind of sediment (rock, cobble, sand, silt, ripple (vertical or at 45 degrees)), shadow or other (typically shipwrecks) parts on images. It is very difficult to discriminate the rock and the the cobble and also the sand and silt. However, it is important for the sedimentologists to discriminate the sand and the silt. The type ``ripple'' can be some ripple of sand or ripple of silt. Hence, with the point of view of the sedimentologists we consider only the three classes of sediment: $C_1$=rock-cobble, $C_2$=sand and $C_3$=silt. And in order to evaluate our decision process, we take the ripple as the fourth class ($C_4$) that is unlearned. 

Each image is cut off in tiles of size 32$\times$32 pixels (about 6.5 meter by 6.5 meter). With these tiles, we keep 3500 tiles of each class with only one kind of sediment in the tile. Hence, our database is made of 4$\times$3500 tiles. We consider 2/3 of them for the learning step (only for the three classes of sediment) and 1/3 of them for the test step (\emph{i.e.} 1167 tiles for each kind of sediment).

In order to classify the tiles of size 32$\times$32 pixels, we first have to extract texture parameters from each tile. Here, we choose the co-occurrence matrices approach \cite{Martin05}. The co-occurrence matrices are calculated by numbering the occurrences of identical gray level of two pixels. Six parameters given by Haralick are calculated: homogeneity, contrast estimation, entropy estimation, the correlation, the directivity, and the uniformity. Concerning these six parameters, we calculate their mean on four directions: 0, 45, 90 and 135 degrees. The problem for co-occurrence matrices is the non-invariance in translation. Typically, this problem can appear in a ripple texture characterization. More features extraction approaches can be used such as the run-lengths matrix, the wavelet transform and the Gabor filters \cite{Martin05}.

We use the {\it libSVM} \cite{Chang01}, and after comparing several kernels, we have retained the radial basis function (with $\gamma=1/6$ where 6 is the dimension of the data) and we take weighting of the error $C=1$ because of the data overlap.

\subsection{Results}

The table \ref{TabSVM} shows the results for the SVM classifier with the strategies one-versus-one and one-versus-rest. We note that there are many errors between the sand ($C_2$) and silt ($C_3$), that are two homogeneous sediments. The ripple ($C_4$), the unlearning class, is more heterogeneous than the sand and silt, this why it is more classified as rock ($C_1$).
\begin{table}[ht]
\centering
  \begin{tabular}{|c|c|c|c||c|c|c|}
    \hline
    & \multicolumn{3}{|c||}{one-vs-one} & \multicolumn{3}{c|}{one-vs-rest}\\
    \hline
    \% & $C_1$ & $C_2$ & $C_3$ & $C_1$ & $C_2$ & $C_3$ \\
    \hline
    $C_1$  & 91.00 & 8.83 & 0.17 & 84.40 & 15.08 & 0.51 \\
    \hline
   $C_2$  & 7.11 & 80.72 & 12.17 & 2.57 & 61.27 & 36.16 \\
    \hline
   $C_3$  & 2.06 & 30.42 & 67.52 & 0.86 & 22.71 & 76.44 \\
    \hline
   $C_4$  & 65.13 & 33.16 & 1.71 & 52.36 & 45.41 & 2.21 \\
    \hline
  \end{tabular}
\caption{Results of the SVM classifier for the both strategies one-versus-one and one-versus-rest.}
\label{TabSVM}
\end{table}
The table \ref{BF_pign} shows the same results, but with the proposed approach based on the belief function theory (presented in section \ref{belief}) with the decision based on the pignistic probability. This approach provides some similar results than the basic versions of the SVM (table \ref{TabSVM}). Note that the strategy one-versus-rest provides more errors between the sand and silt. This can be explained because the data overlap. In the rest of the paper we consider only the one-versus-one strategy.
\begin{table}[ht]
\centering
  \begin{tabular}{|c|c|c|c||c|c|c|}
    \hline
    & \multicolumn{3}{|c||}{one-vs-one} & \multicolumn{3}{c|}{one-vs-rest}\\
    \hline
    \% & $C_1$ & $C_2$ & $C_3$ & $C_1$ & $C_2$ & $C_3$ \\
    \hline
    $C_1$  & 88.00 & 11.91 & 0.09 & 91.51 & 6.18 & 2.31 \\
    \hline
   $C_2$  & 4.80 & 83.29 & 11.91 & 8.83 & 20.90 & 70.27 \\
    \hline
   $C_3$  & 1.20 & 32.05 & 66.75 & 2.06 & 5.23 & 92.71 \\
    \hline
   $C_4$  & 56.38 & 41.99 & 1.63 & 67.52 & 20.57 & 11.91 \\
    \hline
  \end{tabular}
\caption{Results of the SVM classifier with belief function theory for the both strategies one-versus-one and one-versus-rest.}
\label{BF_pign}
\end{table}

\begin{figure}[htb]
\includegraphics[height=6cm]{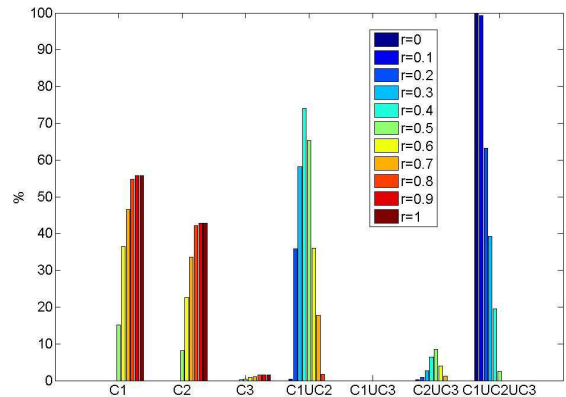}
\vspace{-0.5cm}
\caption{Classification of the class of ripple ($C_4$) with possible decision on union according to $r$.}
\label{VariationR}
\end{figure}

The table \ref{Tab1c1Union} gives the results with possible decision on unions with $r=0.6$. We can see that this kind of cautious decision provides less hard errors (\emph{i.e.} say one kind of sediment instead of another). Of course these results depend on the values of $r$ that provide a more or less cautious decision as we can see on figure~\ref{VariationR}. If we add the possibility of rejection to these results (table~\ref{Tab1c1UnionReject}), we can see that the most of rejected tiles come from the ripple (the unknown class $C_4$). For a given class, the rejected tiles come as a majority from the unions (imprecise data). Of course this rejection does not depend on the $r$ value if we begin by the rejection in our decision process (1-2) (presented in section~\ref{belief}). 

\begin{table*}[!htbp]
\centering
  \begin{tabular}{|c|c|c|c|c|c|c|c|}
    \hline
    \% & $C_1$ & $C_2$ & $C_3$ & $C_1 \cup C_2$ & $C_1 \cup C_3$ & $C_2 \cup C_3$ & $C_1 \cup C_2 \cup C_3$ \\
    \hline
    $C_1$  & 76.69 & 6.86 &  0 & 15.68 &0 & 0.77 & 0 \\
    \hline
   $C_2$  & 0.86 & 50.04 & 4.97 & 11.83 & 0 & 32.30 & 0 \\
    \hline
   $C_3$  & 0.35 & 17.05 & 53.21 & 2.14 & 0 & 27.25 & 0 \\
    \hline
   $C_4$  & 38.99 & 22.62 & 1.12 & 33.16 & 0 & 4.11 & 0 \\
    \hline
  \end{tabular}
\caption{Results with a belief combination with possible decision on unions.}
\label{Tab1c1Union}
\end{table*}

\begin{table*}[!htbp]
\centering
  \begin{tabular}{|c|c|c|c|c|c|c|c|c|}
    \hline
    \% & $C_1$ & $C_2$ & $C_3$ & $C_1 \cup C_2$ & $C_1 \cup C_3$ & $C_2 \cup C_3$ & $C_1 \cup C_2 \cup C_3$ & $C_4$ \\
    \hline
    $C_1$  & 76.69 & 6.26 & 0 & 6.34 & 0 & 0.34 & 0 & 10.37\\
    \hline
   $C_2$  & 0.86 & 48.93 & 4.97 & 1.63 & 0 & 19.45 & 0 & 24.16\\
    \hline
   $C_3$  & 0.34 & 15.42 & 53.22 & 0.34 & 0 & 12.77 & 0 & 17.91\\
    \hline
   $C_4$  & 38.99 & 20.65 & 1.11 & 9.68 & 0 & 1.63 & 1.71 & 27.93\\
    \hline
  \end{tabular}
\caption{Results with a belief combination with possible decision on unions and on the rejected class.}
\label{Tab1c1UnionReject}
\end{table*}

Figure~\ref{VariationR} shows the results of the classification of class of ripple ($C_4$) according to the value of $r$ without possible rejection. Of course when the value of $r$ is weak the data of the three learning classes are classified on the unions. We can distinguished three kinds of work intervals on these data:
\begin{itemize}
\item	$r \in [0 ; 0.3]$: The classifier is too undecided,
\item	$r \in [0.4 ; 0.6]$: the ambiguity between the classes is correctly considered,
\item	$r \in [0.7 ; 1]$: the decision is too hard.
\end{itemize}

According to the application, if we want to privilege the hard decision at the expense of the rejection, we can try to decide first, possibly on the unions and next try to reject only on the unions. In this case we can choose a higher value of $r$. For example with $r=0.8$, we propose a comparison of the decision processes (1-2) with (2-1) given in the table~\ref{Tab1c1_2-1_1-2}. Of course the decision process (2-1) rejects less data, but it is only with the rock (class $C_1$) that we win, and we reject less ripple (class unknown $C_4$). Hence, it seems that the decision process (1-2) is better for this application.

\begin{table*}[ht]
\centering
  \begin{tabular}{|c|c|c|c|c|c|c|c|c|}
    \hline
    \% & $C_1$ & $C_2$ & $C_3$ & $C_1 \cup C_2$ & $C_1 \cup C_3$ & $C_2 \cup C_3$ & $C_1 \cup C_2 \cup C_3$ & $C_4$ \\
    
    \hline
    $C_1$  & 82.86 & 6.77 & 0 & 0 & 0 & 0 & 0 & 10.37\\
    \hline
   $C_2$  & 2.23 & 64.44 & 9.17 & 0 & 0 & 0 & 0 & 24.16\\
    \hline
   $C_3$  & 0.69 & 20.48 & 60.92 & 0 & 0 & 0 & 0 & 17.91\\
    \hline
   $C_4$  & 48.67 & 21.94 & 1.46 & 0 & 0 & 0 & 0 & 27.93\\
    \hline
    \hline
    $C_1$  & 87.75 & 4.20 & 0 & 0.43 & 0 & 0 & 0 & 5.06\\
    \hline
   $C_2$  & 4.54 & 64.44 & 9.17 & 0.34 & 0 & 0 & 0 & 21.51\\
    \hline
   $C_3$  & 1.20 & 20.48 & 60.93 & 0.08 & 0 & 0 & 0 & 17.31\\
    \hline
   $C_4$  & 55.78 & 21.94 & 1.46 & 1.20 & 0 & 0 & 0 & 19.62\\
    \hline
  \end{tabular}
\caption{Results with a belief combination with possible decision on unions and on the rejected class (1-2) and with rejection on the union only (2-1).}
\label{Tab1c1_2-1_1-2}
\end{table*}

Now let's consider the tiles containing more than two kinds of sediment. We still learn the SVM classifier with the same parameter and the one-versus-one strategy on the homogeneous tiles of the three classes rock ($C_1$), sand ($C_2$) and silt ($C_3$) as previously. For the tests, we only take 299 tiles with the classes: $S_1$=tiles with rock and sand, $S_2$=sand and silt, $S_3$=silt and ripple and $S_4$=sand and ripple.

Table~\ref{HeterogensansUnion} presents the obtained results of the SVM classifier with the classical voting combination and a belief combination with pignistic decision and with credibility with reject decision. For the two classes $S_1$ and $S_2$, the tiles contain only learning sediment (rock and sand for $S_1$ and sand and silt for $S_2$). For $S_1$ and $S_2$ the classifiers without reject classify these tiles more in sand. The rejection decreases the errors, but for $S_2$ the rejection is essentially on the sand. The two classes $S_3$ and $S_4$ contain ripple, the unknown class. Here also, we note a confusion with the rock sediment that is an heterogeneous texture like the ripple. The rejection for these two classes works well, because a large part of the tiles classified in rock are rejected and for $S_3$ a large part of tiles classified in sand are also rejected.  
\begin{table}[b]
\centering
  \begin{tabular}{|c|c|c|c||c|c|c||c|c|c|c|}
    \hline
    & \multicolumn{3}{|c||}{vote} & \multicolumn{3}{c||}{pignistic} & \multicolumn{4}{|c|}{with reject} \\
    \hline
    \%& $\!\!C_1\!\!$& $\!\!C_2\!\!$ & $\!\!C_3\!\!$ & $\!\!C_1\!\!$ & $\!\!C_2\!\!$ & $\!\!C_3\!\!$ & $\!\!C_1\!\!$ &$\!\!C_2\!\!$ & $\!\!C_3\!\!$ & $\!\!C_4\!\!$\\
    \hline
    $\!\!\!S_1\! \! \!$ & $\!\!27.4\!\!$ & $\!\!67.2\!\!$ & $\!\!5.4\!\!$ & $\!\!20.4\!\!$ & $\!\!74.9\!\!$ & $\!\!4.6\!\!$ & $\!\!15.4\!\!$ & $\!\!56.5\!\!$ & $\!\!3.3\!\!$ & $\!\!24.8\!\!$\\
    \hline
   $\!\!\!S_2\!\!\!$ &  $\!\!1.3\!\!$ & $\!\!40.5\!\!$ & $\!\!58.2\!\!$ & $\!\!0.3\!\!$ & $\!\!44.8\!\!$ & $\!\!55.9\!\!$ & $\!\!0\!\!$ & $\!\!14.4\!\!$ & $\!\!47.8\!\!$ & $\!\!37.8\!\!$\\
    \hline
    $\!\!\!S_3\! \! \!$ & $\!\!40.1\!\!$ & $\!\!38.1\!\!$ & $\!\!21.8\!\!$ & $\!\!34.1\!\!$ & $\!\!44.8\!\!$ & $\!\!21.1\!\!$ & $\!\!24.4\!\!$ & $\!\!24.1\!\!$ &  $\!\!18.1\!\!$ & $\!\!34.4\!\!$\\
    \hline
    $\!\!\!S_4\! \! \!$ & $\!\!40.8\!\!$ & $\!\!58.2\!\!$ & $\!\!1.0\!\!$ & $\!\!31.8\!\!$ & $\!\!67.2\!\!$ & $\!\!1.0\!\!$ & $\!\!22.7\!\!$ & $\!\!51.2\!\!$ & $\!\!1.0\!\!$ & $\!\!26.1\!\!$\\
    \hline
  \end{tabular}
\caption{Results of the SVM classifier with the classical voting combination and a belief combination with pignistic decision and with credibility with reject decision.}
\label{HeterogensansUnion}
\end{table}

Table~\ref{HeterogenavecUnion} shows the results with possible decision on the union with $r=0.6$, with and without possible rejection. The addition of the possible decision on the union reduces the errors. The rejection is essentially on the tiles classified on the unions, except for $S_2$ (sand and silt) a lot of classified-sand tiles are rejected, maybe because of the learning step.  

\begin{table*}[!htbp]
\centering
  \begin{tabular}{|c|c|c|c|c|c|c|c|c|}
    \hline
    \% & $C_1$ & $C_2$ & $C_3$ & $C_1 \cup C_2$ & $C_1 \cup C_3$ & $C_2 \cup C_3$ & $C_1 \cup C_2 \cup C_3$ & $C_4$ \\
    
    \hline
    $S_1$  & 8.03 & 49.50 & 1.67 & 26.75 & 0 & 14.05 & 0 & -\\
    \hline
   $S_2$  & 0 & 23.08 & 37.46 & 3.34 & 0 & 36.12 & 0 & -\\
    \hline
   $S_3$  & 16.05 & 22.41 & 15.38 & 30.44 & 0 & 15.72 & 0 & -\\
    \hline
   $S_4$  & 15.72 & 47.49 & 0.33 & 31.44 & 0 & 5.02 & 0 & -\\
    \hline
    \hline
    $S_1$  & 8.03 & 48.16 & 1.67 & 9.36 & 0 & 8.03 & 0 & 24.75\\
    \hline
   $S_2$  & 0 & 12.71 & 37.46 & 0 & 0 & 12.04 & 0 & 37.79\\
    \hline
   $S_3$  & 16.05 & 21.40 & 15.38 & 8.70 & 0 & 5.02 & 0 & 33.45\\
    \hline
   $S_4$  & 15.72 & 47.16 & 0.33 & 7.02 & 0 & 3.68 & 0 & 26.09\\
    \hline
  \end{tabular}
\caption{Results with a belief combination with possible decision on unions with and without possible rejection.}
\label{HeterogenavecUnion}
\end{table*}

Hence, for the tiles containing more than one kind of sediments our decision support could help the human experts. Of course, in this case, the evaluation is really difficult. In \cite{Martin06} we have propose confusion matrices taking into account the proportion of each sediment in a tile. 

\section{Conclusions}
We have proposed an original approach based on the belief functions theory for the combination of binary classifiers coming from the SVM with one-versus-one or one-versus-rest strategies. The modelization of the basic belief assignments is proposed directly from the decision functions given by the SVM. These basic belief assignments allow to take correctly into account the principle of the binary classification with SVM by comparison with an hyperplane in linear or nonlinear cases. 

The belief functions theory provides a decision support without necessary deciding an exclusive class. The decision process that we have proposed with possible outliers rejection and with possible decision on the union of classes, is very interesting because it works like the intuitive classification that a human could perform based on the position of support vectors and considering the ambiguity of the classes. This decision support can really help experts for seabed characterization from sonar images. We have seen with the point of view of the sedimentologists that if we only consider the different kinds of sediments (rock, sand and silt), the ambiguity between the sand and the silt is well recognize and the ripple can be partly rejected.

\bibliographystyle{IEEEtran}
% argument is your BibTeX string definitions and bibliography database(s)

% that's all folks
\end{document}